\documentclass[letterpaper, 10 pt, conference]{ieeeconf}
\IEEEoverridecommandlockouts 
\usepackage{graphicx}
\usepackage{setspace}
\usepackage{graphics}
\usepackage{epsfig}
\usepackage{amsmath}
\usepackage{amssymb}
\usepackage{epstopdf}
\usepackage[ruled,vlined,lined,linesnumbered,boxed,commentsnumbered]{algorithm2e}
\usepackage{float}
\usepackage{calc}
\usepackage{url}
\usepackage{graphicx}
\usepackage{relsize}
\usepackage{multirow}
\usepackage{booktabs}
\usepackage{graphicx}
\usepackage{multicol}
\usepackage{tikz}
\usetikzlibrary{shapes,arrows}
\usepackage{subfigure}
\usepackage{xcolor,colortbl}
\usepackage{comment}
\usepackage[breaklinks=true,bookmarks=false]{hyperref}

\newcommand{\myvector}[1]{{\begin{bmatrix} #1 \end{bmatrix}}}

\newcommand{\OrderAlg}[1]{\mathcal{O}({#1})}
\newcommand{\vv}[1]{ \mathbf{#1}} 
\newcommand{\card}[1]{ \vert{#1}\vert}
\newcommand{\Dmh}{d_{\rm MH}}   
\newcommand{\dmh}[2]{ \mathfrak{d}_{#2}^{#1}  } 

\newtheorem{mytheorem}{{\bf \em Theorem}}
\newtheorem{myremark}{{\bf \em Remark}}

\title{\LARGE \bf Learning Binary Features Online from Motion Dynamics for Incremental Loop-Closure Detection and Place Recognition}

\author{Guangcong Zhang$^{1}$, Mason J. Lilly$^{2}$, and Patricio A. Vela$^{1}$%
\thanks{$^{1}$Guangcong Zhang and Patricio A. Vela are with School of Electrical \& Computer Engineering, and Institute of Robotics and Intelligent Machines, Georgia Institute of Technology, North Ave NW, Atlanta, GA 30332, USA. Mason Lilly is with Transylvania University, Lexington, KY 40508, USA.  Mason Lilly's work was done when he was in Georgia Tech during a summer research program.
{\tt\small \{zhanggc, pvela\}@gatech.edu, jmlilly16@transy.edu.}}%
}
\begin{document}
\maketitle
\thispagestyle{empty}
\pagestyle{empty}

\begin{abstract}
This paper proposes a simple yet effective approach to learn visual features online for improving loop-closure detection and place recognition, based on bag-of-words frameworks. The approach learns a codeword in bag-of-words model from a pair of matched features from two consecutive frames, such that the codeword has temporally-derived perspective invariance to camera motion. The learning algorithm is efficient: the binary descriptor is generated from the mean image patch, and the mask is learned based on discriminative projection by minimizing the intra-class distances among the learned feature and the two original features. A codeword for bag-of-words models is generated by packaging the learned descriptor and mask, with a masked Hamming distance defined to measure the distance between two codewords. The geometric properties of the learned codewords are then mathematically justified. In addition, hypothesis constraints are imposed through temporal consistency in matched codewords, which improves precision. The approach, integrated in an incremental bag-of-words system, is validated on multiple benchmark data sets and compared to state-of-the-art methods. Experiments demonstrate improved precision/recall outperforming state of the art with little loss in runtime. 
\end{abstract}

\section{Introduction}
Long-distance visual Simultaneous Localization and Mapping experiences drift that require sophisticated approaches to handle in both uncooperative~\cite{GFSLAM,OOMC} and cooperative cases~\cite{P1,P2,P3}. When the robot revisits a place, the drift can be greatly reduced by imposing geometric constraints in the posterior optimization (e.g. bundle adjustment)~\cite{ORBSLAM}. Identifying when a robot has returned to a previously visited place is referred as loop-closure detection. It plays a key role in visual SLAM systems.

Since~\cite{FabMap} appearance-based methods have become prevalent in visual loop closure detection, due to their decoupling from the location estimate of the robot. A key research in appearance-based methods is what kind of visual features best describe the scene. Along with the progress of feature descriptor designs, various features have been used in loop-closure detection, from the traditional floating point arithmetic based features such as SIFT~\cite{SIFT}, SURF~\cite{SURF}, to the recent binary-encoded features \cite{BRIEF, BRISK, ORB}. In the typical bag-of-words loop-closure system, a feature is extracted from a single frame after it is matched with the previous frame, then potentially used as a codeword in the vocabulary. Such codewords may not be invariant to the perspective transformation from the robot motion.

A common fact in mobile robot applications, arising from the design of appearance-based loop-closure systems, is that the robot often requires similar motion to trigger a loop-closure. The loop-closing image sequence is captured under similar perspective transformations. The key idea in this paper is to learn the codewords by learning feature descriptors invariant to the perspective transformations induced by robot motion. With such codewords, visual features from the same object subjected to perspective distortions are more likely to trigger loop-closure hypotheses and improve the recall. We use binary features due to its overall advantages  demonstrated in loop-closure applications (efficient computation with high precision-recall (PR))~\cite{DBoW, ibuild}. Learning binary features is be done by treating the image patches from a matched pair together with the mean patch as a single class, then optimizing the binary test by minimizing the intra-class distance and maximizing the inter-class distance through Linear Discriminants Analysis (LDA)~\cite{BOLD, LDAFeatLearn1, LDAFeatLearn2}. Furthermore, because a codeword is learned from two consecutive images, its consistent nature implies that if an frame is retrieved as a loop-closure hypothesis, its previous or next frame should also be a hypothesis. Based on this, our algorithm imposes temporal constraints in the hypothesis selection, which further improves the precision. In addition, mathematical analysis shows that the codewords learned by our method have nice geometric properties, which theoretically supports the proposed method.

The major contributions of our paper include:
\begin{itemize}
\item an efficient algorithm based on LDA for learning codewords invariant to perspective transformations from robot motion, involving only matrix additions on image patch and bit-wise operations on binary vectors;
\item theoretical justification for the geometric properties in the learned codeword, demonstrating that the learned codewords can be viewed as ``centroids" in the space induced by the modified Hamming distance;
\item integration into the incremental bag-of-word loop-closure detection system with additional simple hypothesis constraints that demonstrate improved PR with trivial runtime loss on various benchmark data sets.
\end{itemize}

\section{Related Work}
Research effort has sought to improve the pipeline of appearance-based methods, involving: (1) the visual features chosen as codewords; (2) loop-closure retrieval models, e.g. probabilistic model in Fab-Map~\cite{FabMap}, bag of visual words model~\cite{incrementalBoW, DBoW}; (3) data structures for vocabulary storage and search, e.g. Chow-Liu tree~\cite{FabMap}, vocabulary tree for binary features~\cite{DBoW}; and (4) online incremental design in order to get rid of offline training, e.g. IBuILD system~\cite{ibuild}. This review focuses on the visual features used in the loop closure detection, which is highly driven by the development of feature descriptors. Early work in visual descriptors based on floating-point arithmetic, such as SIFT~\cite{SIFT} and SURF~\cite{SURF}, are typically too expensive in computation to fulfill real-time loop-closure detection. The milestone work Fab-map~\cite{FabMap} mitigates this problem by using quantized SURF descriptors, which are encoded in binary vectors. Such an approximation trades precision-recall with runtime. In~\cite{incrementalBoW}, the authors use raw SIFT descriptors with an additional feature space of local color histogram. With a tree-structure vocabulary in their bag-of-word model, high frame rates are attained with the feature retrieval of logarithmic-time complexity in codeword number. The work in~\cite{randomTreeLoopClosure} attains real-time performance by using compact randomized tree signatures. With the developments of binary descriptors such as BRIEF~\cite{BRIEF}, BRISK~\cite{BRISK}, ORB~\cite{ORB}, binary features have been widely adopted for loop-closure detection (often with bag-of-word models) due to their fast computation and comparable precision/recall to SIFT and SURF. The new standard, a binary features based bag-of-word system, is presented in~\cite{DBoW}. The IBuILD system~\cite{ibuild} further improves the binary bag-of-word recipe by designing an online incremental system without the need for prior feature training. Our system improves upon IBuILD.

Beside standard feature descriptors, related research has sought to improve descriptors through additional learning processes. These learning approaches include LDA~\cite{LDAFeatLearn1, LDAFeatLearn2, LDAFeatLearn3, BOLD}, boosting~\cite{binBoost}, Principal Component Analysis (PCA)~\cite{DAISY}, Domain-Size Pooling~\cite{DSP_SIFT}, etc. Our work is inspired by~\cite{BOLD}, in which LDA is applied by learning a mask \textbf{locally} to minimize the intra-class distance of binary descriptors. The algorithm synthesizes samples from each image by rotating the patch, and treats the original patch as a pivot patch along with the axillary samples to form a single class for LDA. To contrast, our method first uses the synthesized patch as the pivot patch and the original patches as auxillary patches. More importantly, instead of learning the invariance to some heuristic rotation transformations, our method learns the transformation invariance for the actual robot motion. 

\section{Learning Binary Codeword Invariant to Frame-by-Frame Motion Dynamics}
This section first presents basics about binary feature descriptors and the LDA method for optimizing the descriptor by learning a mask. Then it details the proposed algorithm for learning features from motion dynamics. The geometric properties will be discussed in detail. Out notation uses boldface for a vector or matrix (e.g., $\vv{x}$), and normal font for a scalar binary or real value (e.g., $x_i$).

\subsection{Binary Descriptors and Intra-class Distance}

\subsubsection{Binary Descriptors for a visual feature}~\\
Binary descriptors~\cite{BRIEF, BRISK, ORB} follow the same basic formulation. Given an image intensity patch $I$, a binary descriptor is encoded as a binary vector $\vv{x}$, composed of $L$ bits $x_i \in \mathbb{B}$ (typically $L=512$). Often after a smoothing operation, each bit in $\vv{x}$ is generated by binary tests $\{[\vv{a_i}, \vv{b_i}]\}_{i=1}^L$
\begin{equation}
\label{eq:binary_tests}
x_i = \left\{
	\begin{array}{ll}
		1  & \text{if } I(\vv{a_i}) < I(\vv{b_i}) \\
		0  & \text{otherwise}
	\end{array}\right.
,\quad \forall i = 1 \dots L
\end{equation}
where each $\vv{a_i}=\myvector{u_{a_i}, v_{a_i}}^\top$ (and similarly $\vv{b_i}$) is a pixel position. 
The binary tests pattern are usually generated offline by training on large data sets. Here, we use the BOLD binary test pattern~\cite{BOLD}.

\subsubsection{Intra-class distance}~\\
The distance between two binary descriptors is measured by the Hamming distance $d_{\rm H}$ with bit-xor operation $\oplus$: 
\begin{equation} 
	\begin{split}
		d_{\rm H} &\left(\vv{x}^{(k)}, \vv{x}^{(k')} \right) =\vv{x}^{(k)} \oplus \vv{x}^{(k')} \\ 
		&= \sum_{i=1}^L {x}^{(k)}_l  \oplus {x}_l^{(k')} = \sum_{i=1}^L \left( {x}^{(k)}_l  - {x}_l^{(k')} \right)^2
	\end{split}
\end{equation}

For a set of image patches $\{I^{(k)}\}$ from the same class and the corresponding binary descriptors $\{\vv{x}^{(k)}\}$, the expected intra-class distance is:

\begin{align} \label{eq:intraclass_distance}
    \nonumber
	\mathbb{E} & \left[d_{\rm H} (\{\vv{x}^{(k)}\})\right] = \frac{1}{L} \sum_{l=1}^L \mathbb{E} \left[d_{\rm H} (\{x_l^{(k)}\})\right] \\ \nonumber
	& = \frac{1}{LK^2} \sum_{l=1}^L  \sum_{k=1}^K \sum_{k'=1}^K d_{\rm H} (x_l^{(k)}, x_l^{(k')}) \\ \nonumber
	& = \frac{1}{LK^2} \sum_{l=1}^L  \sum_{k=1}^K \sum_{k'=1}^K \left( {x}^{(k)}_l  - {x}_l^{(k')} \right)^2 \\ \nonumber
	& = \frac{1}{LK^2} \sum_{l=1}^L \left( 2\sum_{k=1}^K \sum_{k=1}^K \left(x_l^{(k)}\right)^2 - 2\sum_{k=1}^K \sum_{k'=1}^K x_l^{(k)}x_l^{(k')} \right)\\ 
	& = \frac{1}{L} \sum_{l=1}^L  2\mathbb{E}\left[x_l^2\right] -2\mathbb{E}\left[x_l\right]^2
\end{align}

\subsection{Learning Codewords from Motion Dynamics}

\subsubsection{Minimizing intra-class distances with binary masks}
Eq.~\ref{eq:intraclass_distance} shows that minimizing the intra-class distance can be done by masking the binary coordinates with high variance, effectively projecting out the highly variable coordinates. We will therefore package learned codewords into a feature ensemble consisting of a feature descriptor and a binary mask $\vv{D} = \{\vv{x}, \vv{y}\} \in \mathcal{D}_{\rm MH}$, where the mask $\vv{y}$ is defined
\begin{equation}
  \begin{split}
	  y_i \cong \left\{
	  	\begin{array}{ll}
	  		1  & \text{if}~(\land_{k} x_i^{(k)}=1) \lor
              (\land_{k} x_i^{(k)}=0) \\
	  		0  & \text{otherwise}
	  	\end{array}\right.
  \end{split}
\end{equation}
for each coordinate $i \in \left\{1 \dots L\right\}$.
The distance between two feature ensembles, with non-zero masks, is defined to
be ``masked Hamming distance" $\Dmh$:
%
\begin{align}
    \nonumber
	\Dmh(\vv{D}_1, \vv{D}_2) &= \frac{\card{\vv{y}_2} \card{\vv{x}_1\oplus \vv{x}_2 \cap \vv{y}_1} + \card{\vv{y}_1} \card{\vv{x}_1\oplus \vv{x}_2 \cap \vv{y}_2 }}{\card{\vv{y}_1}+\card{\vv{y}_2}} \\ 
	&= \frac{\card{\vv{y}_2}}{\card{\vv{y}_1}+\card{\vv{y}_2}} \dmh{1}{2} + \frac{\card{\vv{y}_1}}{\card{\vv{y}_1}+\card{\vv{y}_2}} \dmh{2}{1} 
\label{eq:maskedHamming}
\end{align}
where $\card{\cdot}$ is the number of 1s in a binary vector; $\dmh{i}{j} \triangleq \card{\vv{x}_i\oplus \vv{x}_j \cap \vv{y}_i}$ and $\dmh{i}{j} \neq\dmh{j}{i}$. 

The distance metric defined in Eq.~\ref{eq:maskedHamming} has two significant properties: (1) $\Dmh(\vv{D}_1, \vv{D}_2)\in [0, L]$. The lower bound is straightforward, and the upper bound is simply given by $\Dmh(\vv{D}_1, \vv{D}_2) \leq \frac{2\card{\vv{y}_1} \card{\vv{y}_2}}{\card{\vv{y}_1}+\card{\vv{y}_2}} \leq \frac{2\card{\vv{y}_1} \card{\vv{y}_2}}{2\sqrt{\card{\vv{y}_1}\card{\vv{y}_2}}} = \sqrt{\card{\vv{y}_1}\card{\vv{y}_2}} \leq L$.  (2) When $\vv{y}_1, \vv{y}_2$ are all 1s, then $\Dmh(\vv{D}_1, \vv{D}_2) \equiv d_{\rm H}(\vv{x}_1, \vv{x}_2)$, which means the masked Hamming distance defined also accommodates the Hamming distance. The masked Hamming distance is not a metric since the coincidence axiom and the triangular inequality do not hold.

\subsubsection{Learning codewords invariant to cross-frame motion}
\begin{figure}[t!]
  \centering
  \includegraphics[width=0.99\linewidth]{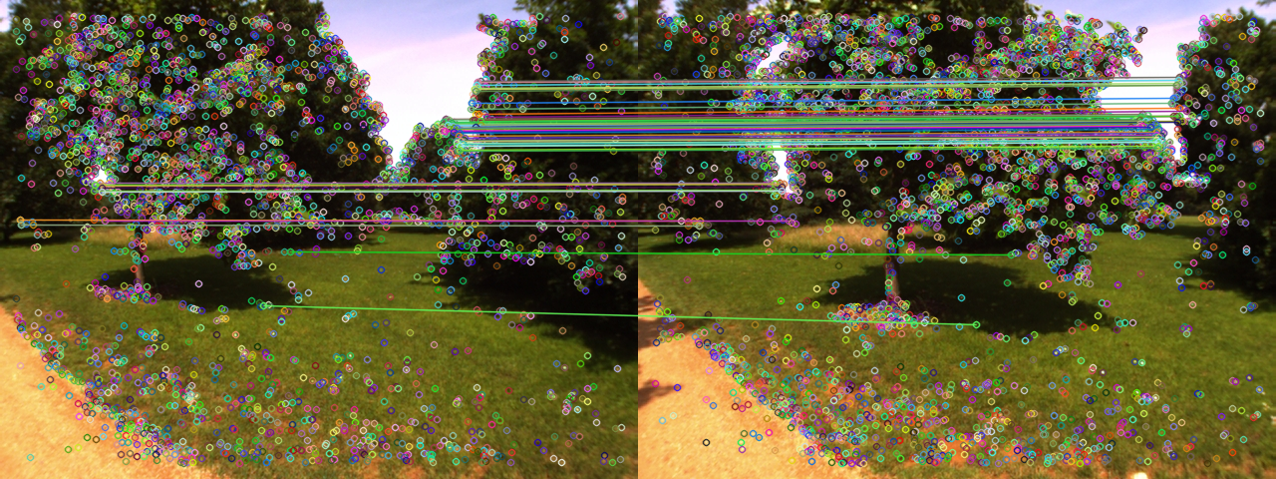}
  \caption{For each pair of matched frame, we extract a $48\times 48$ patch $I_1$ from the previous frame, and $I_2$ from the current frame. These two will be used to learn a codeword.\label{fig:frameMatch}}
\end{figure}
\begin{figure}[t!]
  \centering
  \includegraphics[width=0.9\linewidth]{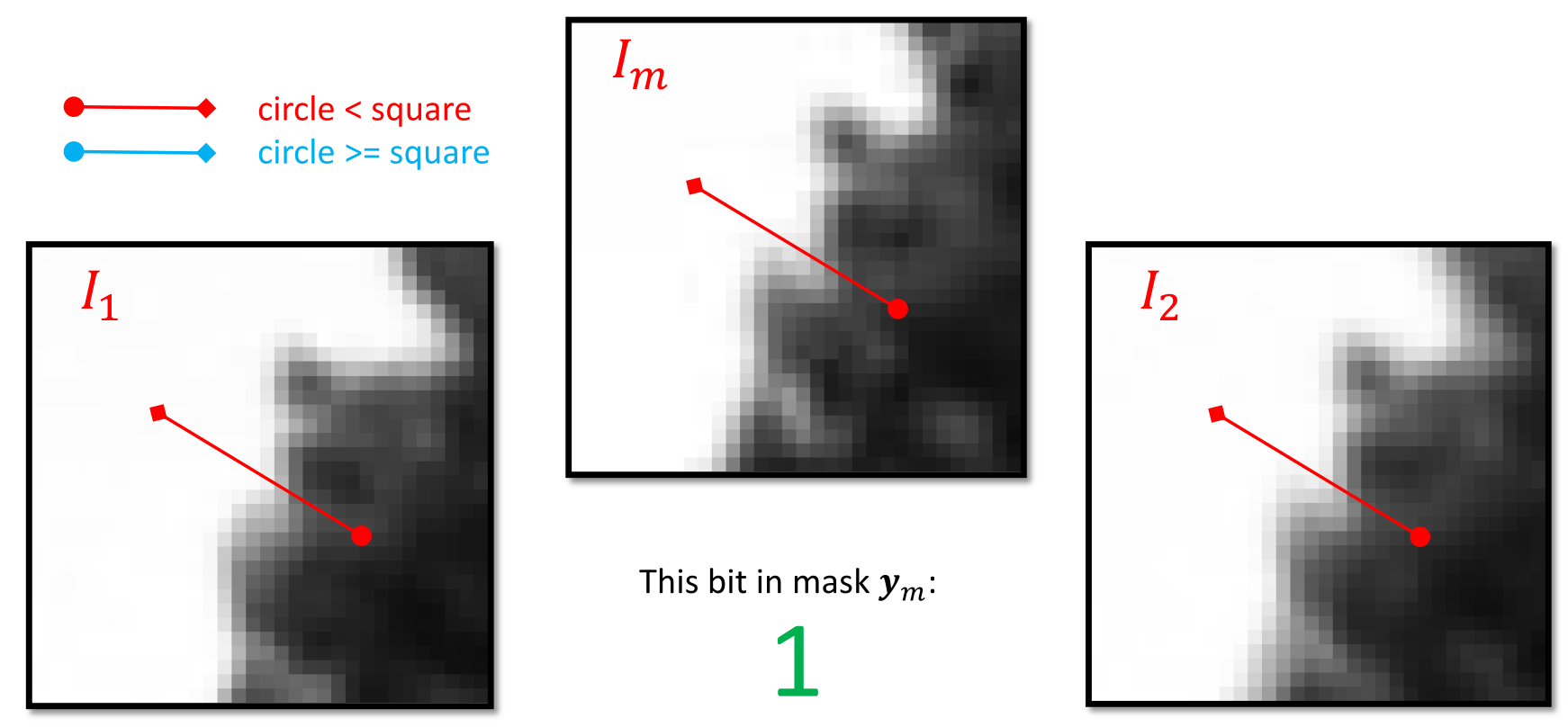}
  \noindent\rule{8.5cm}{1.5pt}
  \includegraphics[width=0.9\linewidth]{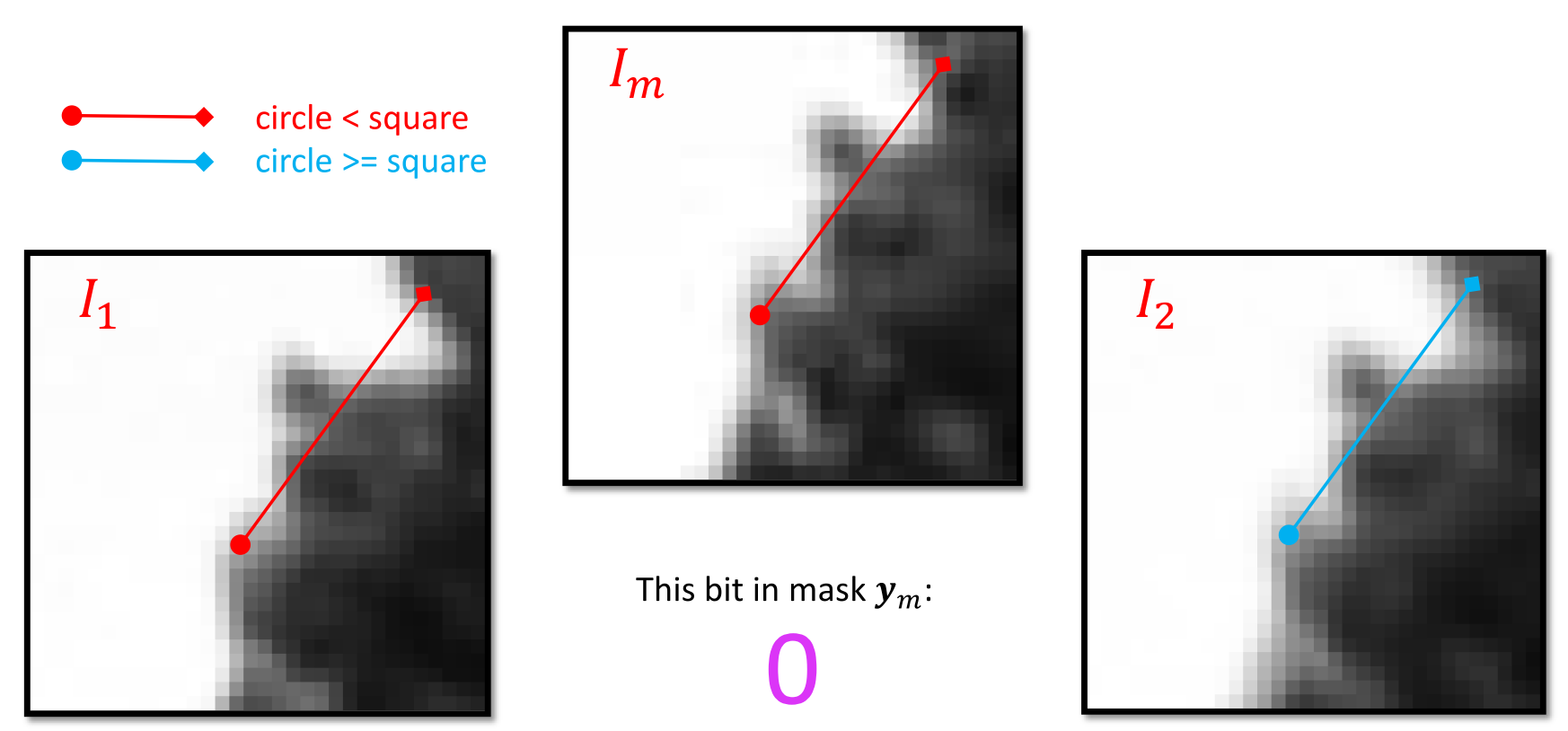}
  \caption{Illustration of binary tests and mask learning for $I_m$ from $I_1, I_2$.\label{fig:ImTests}}
\end{figure}
\begin{figure}[t!]
	\centering
	\includegraphics[width=0.75\linewidth]{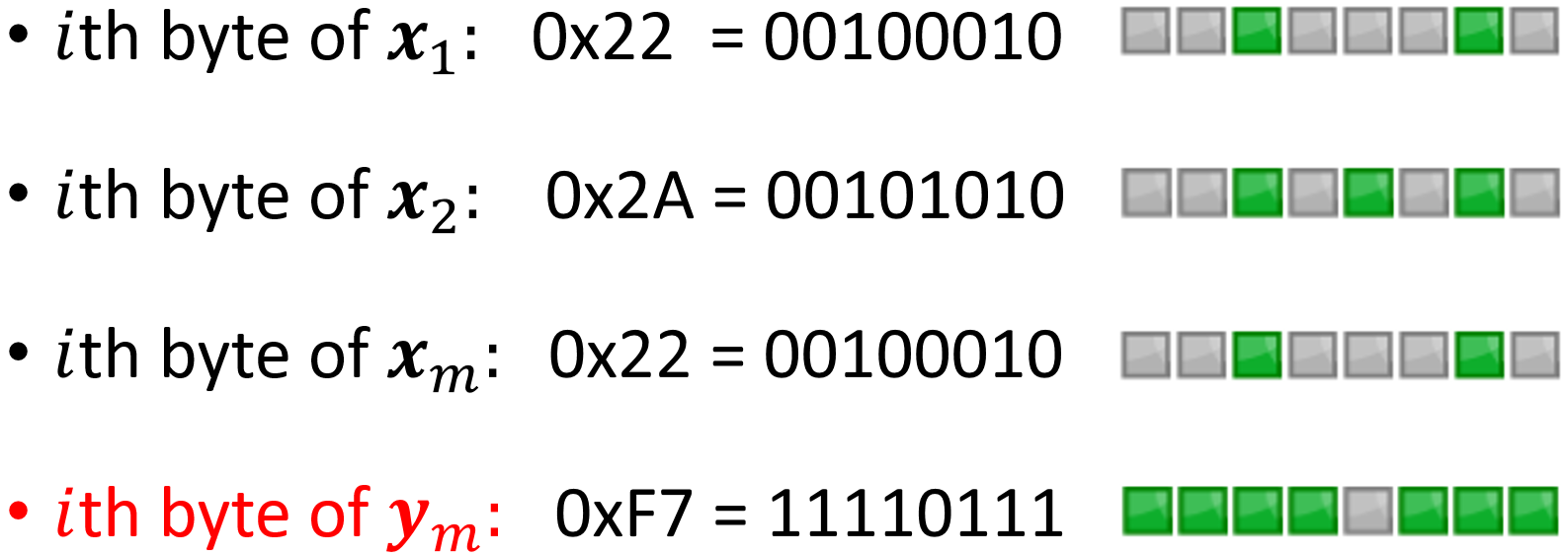}
	\caption{Codeword learning from the perspective of bit-wise operations.\label{fig:YmIllu}}
\end{figure}

Given a pair of matched image patches $I_1, I_2$ from two consecutive frames (as depicted in Fig.~\ref{fig:frameMatch}), the mean patch $I_m$ is used to generate the codeword $\vv{D}_m = \{\vv{x}_m, \vv{y}_m\}$,
\begin{equation}
	I_m = \frac12\left(I_1 + I_2\right).
\end{equation}
$I_m$ has the structural information of $I_1, I_2$. 
Binary tests on $I_1$, $I_2$, and $I_m$, per Eq.~\ref{eq:binary_tests}, 
generate the binary vectors $\vv{x}_1$, $\vv{x}_2$, and $\vv{x}_m$.
The masks $\vv{y}_1$ and $\vv{y}_2$ are set to all 1s.
The mask $\vv{y}_m$ is computed by minimizing the intra-class distance 
among $\{I_m, I_1, I_2\}$:
\begin{equation}
\label{eq:CodewordMask}
  \begin{split}
	  y_{m,i} = \left\{
	  	\begin{array}{ll}
	  		1  & \text{if}~\bigcap_{k\in \{1, 2, m\}} \overline{\left(I_k(\vv{a_i}) < I_k(\vv{b_i})\right)}=1   \\
	  		   &  \ \text{{or}}~\bigcap_{k\in \{1, 2, m\}} \left(I_k(\vv{a_i}) < I_k(\vv{b_i})\right)=1 \\
	  		0  & \text{otherwise}
	  	\end{array}\right.
  \end{split}
\end{equation}
The algorithm definined the mean codeword is summarized in Algorithm~\ref{alg:Codeword_generation}\;.
Fig.~\ref{fig:ImTests} depicts two binary tests on $\{I_m, I_1, I_2\}$. If non-zero variance exists in the same test across three patches, the corresponding dimension in $\vv{x}_m$ will be masked out. Fig.~\ref{fig:YmIllu} illustrates the process in terms of binary vector expressions.  
\begin{algorithm}[t]
 \KwData{ $I_1, I_2 \in \mathbb{R}^{M\times N}$ -- imaged patches from a pair of matched features in two consecutive frames; $\{[\vv{a_i}, \vv{b_i}]\}$ -- lists of binary tests positions }

 \KwResult{ $\vv{D}_m = \{\vv{x}_m, \vv{y}_m\}$ -- Codeword of length $L$ invariant to perspective transformation between $I_1, I_2$}
 
	$I_m \gets \frac 12 (I_1 + I_2)$
	\tcp*[r]{\small $\OrderAlg{MN}$}
	
	$\vv{x}_m \gets$ \bf{BinaryTests} $\left(I_m, \{[\vv{a_i}, \vv{b_i}]\} \right)$
	\tcp*[r]{\small Eq.\ref{eq:binary_tests},$\OrderAlg{L}$}
	
	$\vv{y}_m \gets$ \bf{MaskLearning} $\left(\{I_m, I_1, I_2\} \right)$
		\tcp*[r]{\small Eq.\ref{eq:CodewordMask}$,\OrderAlg{L}$}
 \caption{\small \bf Learning codewords from motion dynamics. \label{alg:Codeword_generation}}
\end{algorithm}

\subsection{Geometric Properties of Learned Codewords}
Relative to the source codewords $\vv{D}_1= \{\vv{x}_1, \vv{y}_1\}$ and 
$\vv{D}_2 = \{\vv{x}_2, \vv{y}_2\}$ both  in $\mathcal{D}_{\rm MH}$,
the learned mean codewords have the following two nice geometric properties in the space $\mathcal{D}_{\rm MH}$ induced by $\Dmh$ (note that $\forall ~\vv{D}_k \in \mathcal{D}_{\rm MH},~\card{\vv{y}_k}\neq 0$): 

\begin{enumerate}
\item $\vv{D}_m$ can be viewed as the \textbf{topological centroid} of $\vv{D}_1$ and $\vv{D}_2$;
\item Given any other point $\vv{D}_k \in \mathcal{D}_{\rm MH}$, $\vv{D}_m$ \textbf{preserves the localities} between $\vv{D}_k$ and $\vv{D}_m$, and between $\vv{D}_k$ and $\vv{D}_1, \vv{D}_2$.
\end{enumerate}
These properties are shown to hold in Theorems \ref{The:topoCentroid} and \ref{The:localityPreserve}\;.

\begin{mytheorem}
\label{The:topoCentroid}
Let $\vv{D}_m$ be the codeword generated from $\vv{D}_1$ and $\vv{D}_2$, then 
\begin{equation}
\begin{split}
	&\Dmh(\vv{D}_m, \vv{D}_1) \leq \Dmh(\vv{D}_1, \vv{D}_2) \quad \text{and}\\
    &\Dmh(\vv{D}_m, \vv{D}_2) \leq \Dmh(\vv{D}_1, \vv{D}_2)
\end{split}
\end{equation}
\end{mytheorem}
\vspace{5pt}
\begin{proof}
Here, we prove the first inequality. From Eq.~\ref{eq:CodewordMask}, $\dmh{m}{1} = 0$. Also, because $I_m$ is the mean patch of $I_1$ and $I_2$, $\card{\vv{x}_m\oplus \vv{x}_1} \leq \card{\vv{x}_2\oplus \vv{x}_1}$, which further implies $\dmh{1}{m} = \card{(\vv{x}_m\oplus \vv{x}_1) \cap \vv{y}_1} \leq \card{(\vv{x}_2\oplus \vv{x}_1) \cap \vv{y}_1} = \dmh{1}{2}$. In addition, $\vv{y}_1 = \vv{y}_2 \Longrightarrow \dmh{2}{1} = \dmh{1}{2}$.
Therefore,
%
\begin{align}
	\Dmh(\vv{D}_m, \vv{D}_1)& = \frac{\card{\vv{y}_m}}{\card{\vv{y}_1}+\card{\vv{y}_m}} \dmh{1}{m} + \frac{\card{\vv{y}_1}}{\card{\vv{y}_1}+\card{\vv{y}_m}} \dmh{m}{1}  \\ \nonumber
	&= \frac{\card{\vv{y}_m}}{\card{\vv{y}_1}+\card{\vv{y}_m}} \dmh{1}{m}  \\ \nonumber
	&\leq \frac{\card{\vv{y}_m}}{\card{\vv{y}_1}+\card{\vv{y}_m}} \dmh{1}{2} \\ \nonumber
	&\leq \dmh{1}{2} \\ \nonumber
	& = \frac{\card{\vv{y}_2}}{\card{\vv{y}_1}+\card{\vv{y}_2}} \dmh{1}{2} + \frac{\card{\vv{y}_1}}{\card{\vv{y}_1}+\card{\vv{y}_2}} \dmh{1}{2} \\ \nonumber
	& = \frac{\card{\vv{y}_2}}{\card{\vv{y}_1}+\card{\vv{y}_2}} \dmh{1}{2} + \frac{\card{\vv{y}_1}}{\card{\vv{y}_1}+\card{\vv{y}_2}} \dmh{2}{1} \\ \nonumber
	& = \Dmh(\vv{D}_1, \vv{D}_2)
\end{align}
Likewise, $\Dmh(\vv{D}_m, \vv{D}_2) \leq \Dmh(\vv{D}_1, \vv{D}_2)$ also holds.
\end{proof}
\begin{figure}[t!]
	\centering
	\includegraphics[width=0.5\linewidth]{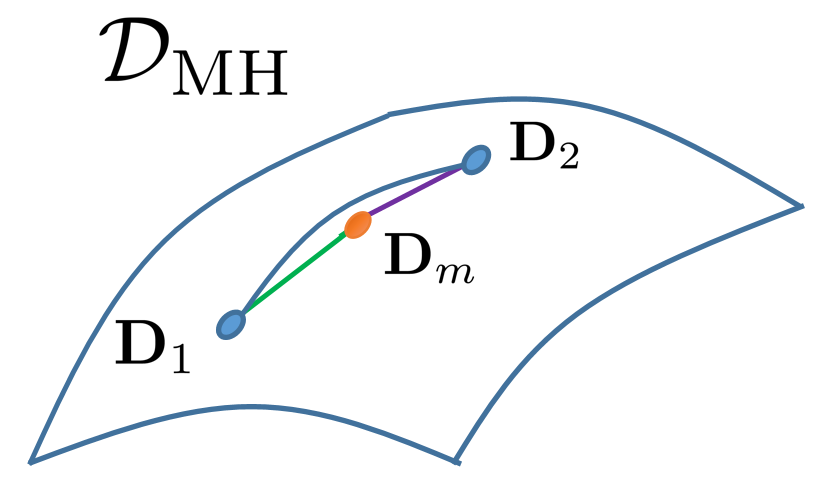}
	\caption{Topologically $\vv{D}_m$ is centroid of $\vv{D}_1$ and $\vv{D}_2$ in $\mathcal{D}_{\rm MH}$.\label{fig:topoCentroid}}
\end{figure}

\vspace{10pt}
\newcommand{\cardY}[1]{ \vert{\mathbf{y}_{#1}\vert}}
\newcommand{\DmhD}[2]{\Dmh(\vv{D}_{#1}, \vv{D}_{#2})}
\newcommand{\DmhWholeEq}[2]{\frac{\card{\vv{y}_{#2}}}{\card{\vv{y}_{#1}}+\card{\vv{y}_{#2}}} \dmh{{#1}}{{#2}} + \frac{\card{\vv{y}_{#1}}}{\card{\vv{y}_{#1}}+\card{\vv{y}_{#2}}} \dmh{#2}{{#1}} }
\newcommand{\DmhWholeEqShort}[2]{\frac{\card{\vv{y}_{#2}} \dmh{{#1}}{{#2}} }{\card{\vv{y}_{#1}}+\card{\vv{y}_{#2}}}  + \frac{\card{\vv{y}_{#1}} \dmh{#2}{{#1}} }{\card{\vv{y}_{#1}}+\card{\vv{y}_{#2}}}}
\newcommand{\DmhWholeEqFirstL}[2]{\frac{\card{\vv{y}_{#2}} \dmh{{#1}}{{#2}}}{L+\card{\vv{y}_{#2}}}  + \frac{L \dmh{#2}{{#1}} }{L+\card{\vv{y}_{#2}}}}
\begin{mytheorem}
\label{The:localityPreserve}
Let $\vv{D}_m$ be the codeword generated from $D_1$ and $D_2$, then $\forall ~\vv{D}_k \in \mathcal{D}_{\rm MH}$, the following inequality holds
\begin{equation}
	\Dmh(\vv{D}_k, \vv{D}_1) + \Dmh(\vv{D}_k, \vv{D}_2) \geq \lambda \Dmh(\vv{D}_k, \vv{D}_m),
\end{equation}
where
\begin{equation}
	\lambda = \left(\frac{\cardY{k}}{L+\cardY{k}} \right) \left[ 1+ \frac{\min(\cardY{m}, \cardY{k})}{\max(\cardY{m}, \cardY{k})} \right] \in [0,1]
\end{equation}
\end{mytheorem}
\vspace{10pt}

\begin{table}[t]
\footnotesize
\centering
\caption{Distances of codewords with one bit (Case\#1). For simplicity we use $\Dmh(i,j)$ to denote $\DmhD{i}{j}$.}
\label{Tab:distOneBitCase1}
	\begin{tabular}{cccc|ccc}
	\hline
	$\vv{D}_1$ & $\vv{D}_2$ & $\vv{D}_m$ & $\vv{D}_k$ & $\Dmh(1,k)$ & $\Dmh(2,k)$ & $\Dmh(m,k)$ \\ \hline
	(1,1) & (1,1) & (1,1) & (1,1) & 0 & 0 & 0 \\
	(1,1) & (1,1) & (1,1) & (1,0) & 0 & 0 & 0 \\
	(1,1) & (1,1) & (1,1) & (0,1) & 1 & 1 & 1 \\
	(1,1) & (1,1) & (1,1) & (0,0) & 1 & 1 & 1 \\ \hline
	\end{tabular}
\end{table}
\begin{table}
\footnotesize
\centering
\caption{Distances of codewords with one bit (Case\#2).}
\label{Tab:distOneBitCase2}
	\begin{tabular}{cccc|ccc}
	\hline
	$\vv{D}_1$ & $\vv{D}_2$ & $\vv{D}_m$ & $\vv{D}_k$ & $\Dmh(1,k)$ & $\Dmh(2,k)$ & $\Dmh(m,k)$ \\ \hline
	(1,1) & (0,1) & (0,0) & (1,1) & 0 & 1 & 1 \\
	(1,1) & (0,1) & (0,0) & (1,0) & 0 & 1 & 0 \\
	(1,1) & (0,1) & (0,0) & (0,1) & 1 & 0 & 0 \\
	(1,1) & (0,1) & (0,0) & (0,0) & 1 & 0 & 0 \\ \hline
	\end{tabular}
\end{table}

\begin{proof}
First let's consider the codewords with only one bit. There are two cases:
\begin{itemize}
	\item[] {\bf Case 1:} If $\vv{x}_{1}$ and $\vv{x}_{2}$ have the same value, WLOG assume this value is $1$, then $\vv{x}_{m} = \vv{y}_{m} = 1$. Since $\vv{x}_k,\vv{y}_{k} \in \{0, 1\}$, there are four situations as listed in Table~\ref{Tab:distOneBitCase1}.
	\item[] {\bf Case 2:} If $\vv{x}_{1}$ and $\vv{x}_{2}$ have the different value, WLOG assume $\vv{x}_1=1, \vv{x}_2=0$, then $\vv{y}_{m} = 0$ and $\vv{x}_{m}$ can be $1$ or $0$. Due to the symmetry, we only need to consider either one of these situations. Let $\vv{x}_{m}=0$. There are again four situations listed in Table~\ref{Tab:distOneBitCase2}.
\end{itemize}

In sum, in any one-dimensional case it always holds that
\begin{equation}
\begin{split}
	\Dmh(m,k) &\leq \Dmh(1,k) + \Dmh(2,k) \\
	\Longleftrightarrow  \dmh{m}{k} + \dmh{k}{m} & \leq \dmh{1}{k} + \dmh{k}{1} + \dmh{2}{k} + \dmh{k}{2}
\end{split}
\end{equation}
For any descriptor with $L$ dimensions, $\dmh{m}{k} + \dmh{k}{m} \leq \dmh{1}{k} + \dmh{k}{1} + \dmh{2}{k} + \dmh{k}{2}$ still holds, because $\dmh{i}{j}$ is simply a summation over all dimensions without weighting.

Also considering $\card{\vv{y}_k} \leq \card{\vv{y}_1} \equiv \card{\vv{y}_2} = L$, we have
%
\begin{align}
	&\DmhD{1}{k} + \DmhD{2}{k} \\ \nonumber
	&=\DmhWholeEqShort{1}{k} + \DmhWholeEqShort{2}{k} \\  \nonumber
	&=\DmhWholeEqFirstL{1}{k} + \DmhWholeEqFirstL{2}{k} \\ \nonumber
	&\geq \frac{\card{\vv{y}_k}}{L+\card{\vv{y}_k}} \left(\dmh{1}{k} + \dmh{k}{1} + \dmh{2}{k} + \dmh{k}{2} \right) \\ \nonumber
	&\geq \frac{\card{\vv{y}_k}}{L+\card{\vv{y}_k}} \left( \dmh{m}{k} + \dmh{k}{m}  \right) \\ \nonumber
	&= \frac{\card{\vv{y}_k}}{L+\card{\vv{y}_k}} \cdot \frac{\card{\vv{y}_k} + \card{\vv{y}_m}}{\max (\card{\vv{y}_k}, \card{\vv{y}_m})}  \\ \nonumber
	&~~~~~~~~~~~~~~\cdot \left( \frac{\max (\card{\vv{y}_k}, \card{\vv{y}_m})}{\card{\vv{y}_k} + \card{\vv{y}_m}} \dmh{m}{k}+ \frac{\max (\card{\vv{y}_k}, \card{\vv{y}_m})}{\card{\vv{y}_k} + \card{\vv{y}_m}}\dmh{k}{m}  \right) \\ \nonumber
	&\geq \frac{\card{\vv{y}_k}}{L+\card{\vv{y}_k}} \cdot \frac{\card{\vv{y}_k} + \card{\vv{y}_m}}{\max (\card{\vv{y}_k}, \card{\vv{y}_m})}  \\ \nonumber
	&~~~~~~~~~~~~~~\cdot \left( \frac{\card{\vv{y}_m}}{\card{\vv{y}_k} + \card{\vv{y}_m}} \dmh{m}{k}+ \frac{\card{\vv{y}_k}}{\card{\vv{y}_k} + \card{\vv{y}_m}}\dmh{k}{m} \right) \\ \nonumber
	&= \left(\frac{\cardY{k}}{L+\cardY{k}} \right) \left[ 1+ \frac{\min(\cardY{m}, \cardY{k})}{\max(\cardY{m}, \cardY{k})} \right] \Dmh(\vv{D}_k, \vv{D}_m)\\ \nonumber
	&\triangleq \lambda\cdot \Dmh(\vv{D}_k, \vv{D}_m) 
\end{align}
It is easy to see that $\lambda > 0$ ($\lambda\neq 0$ because $\cardY{k} \neq 0$). On the other hand, $\lambda = \left(\frac{1}{L/\cardY{k}+1} \right) \left[ 1+ \frac{\min(\cardY{m}, \cardY{k})}{\max(\cardY{m}, \cardY{k})} \right] \leq \left(\frac{1}{L/L +1} \right) \left[ 1+ \frac{\max(\cardY{m}, \cardY{k})}{\max(\cardY{m}, \cardY{k})} \right] = 1$. Thus, $\lambda \in [0, 1]$.
\end{proof}

\vspace{7pt}
\begin{myremark}
Theorem~\ref{The:topoCentroid} shows that $\vv{D}_m$ can be seen as topological centroid of $\vv{D}_1$ and $\vv{D}_2$, as depicted in Fig.~\ref{fig:topoCentroid}.
\end{myremark}
\begin{myremark}
The limits $\lambda \in [0, 1]$ in Theorem~\ref{The:localityPreserve} are conservative since they do not factor that $\vv{D}_1, \vv{D}_2$ are matched features. In reality, $\vv{x}_1 \approx \vv{x}_2 \Longrightarrow \card{\vv{y}_m} \approx L$. In practice, $\vv{D}_k$ will also be generated from a matched pair.  Therefore, $\card{\vv{y}_k} \approx L \approx \card{\vv{y}_m}$ also, to conclude $\lambda \approx \frac{1}{2}(1 + 1) = 1$.
\end{myremark}
\begin{myremark}
Intuitively, Theorem~\ref{The:localityPreserve} shows that the locality is preserved by using $\vv{D}_m$ as a proxy. The theorem can be interpreted as ``{\em if $\vv{D}_k$ is far away from $\vv{D}_m$, then $\vv{D}_k$ is also far away from $\vv{D}_1$ and (or) $\vv{D}_2$}". This is important to loop-closure application: if the matching between two codewords $\vv{D}_k$ and $\vv{D}_m$ is rejected and no loop-closure hypothesis is triggered, then a real loop-closure is not likely to exist because $\vv{D}_k$ must be unable to matched with the original features $\vv{D}_1, \vv{D}_2$ up to the factor $\lambda$.
\end{myremark}

\subsection{Temporal Constraints on Loop-closure Hypotheses}
\label{subsec:temporal_constraints}

\begin{figure}[t!]
  \centering
  
  \includegraphics[width=0.85\linewidth]{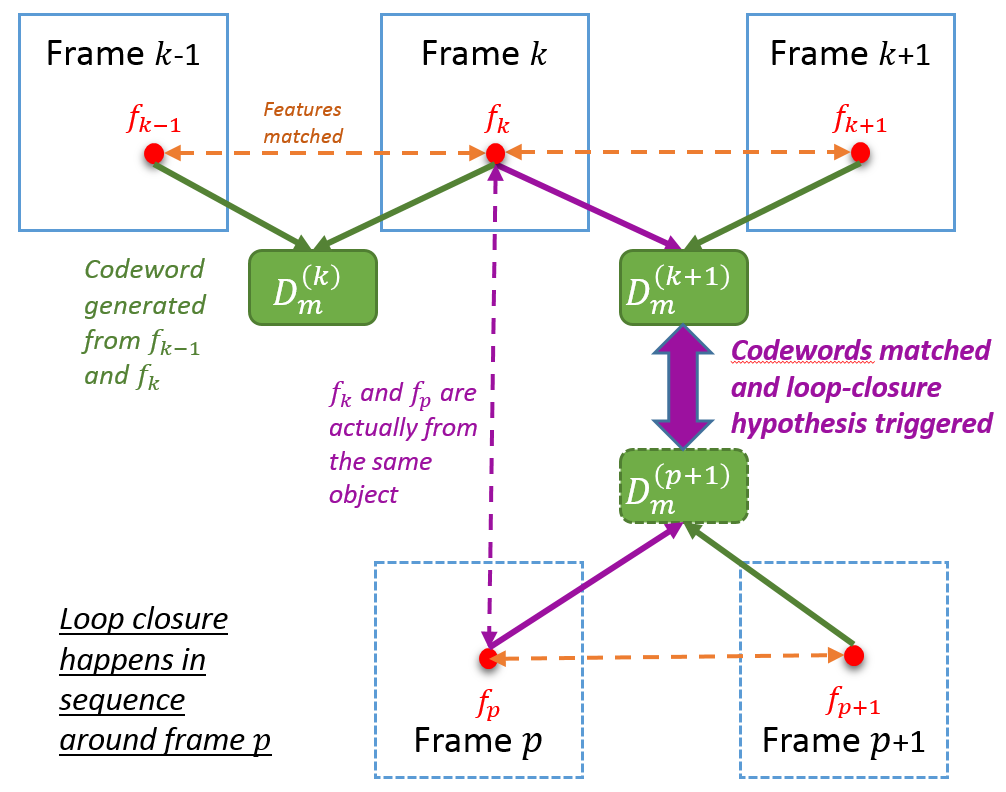}
   \noindent\rule{8.5cm}{1.5pt}
   \vspace{5pt}
   
  \includegraphics[width=0.85\linewidth]{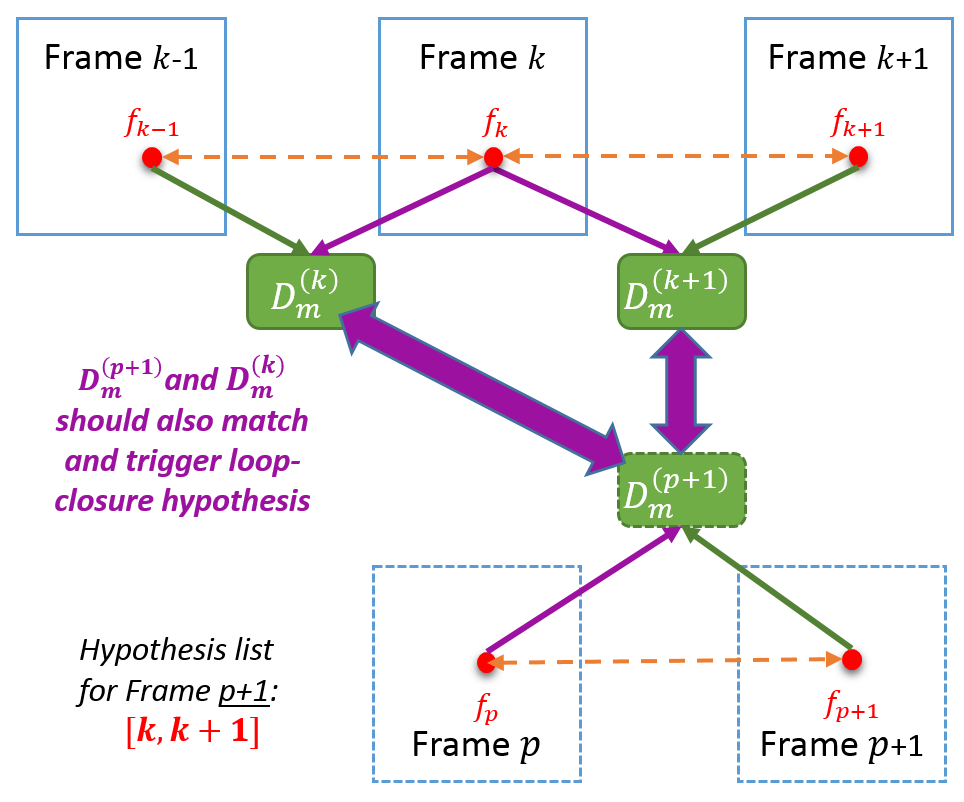}
 
  \caption{Illustration of intrinsic temporal constraint on the loop-closure hypotheses. \label{fig:lcTemporalConstraint}}
\end{figure}

Here we discuss a technique for improving the detection precision, which comes
naturally from the learning process of a codeword.
As illustrated in Fig.~\ref{fig:lcTemporalConstraint}, a loop-closure happens around frame $p$ which revisits the same place captured previously by sequence around frame $k$. A loop-closure hypothesis closing frame $p+1$ with frame $k+1$ is triggered in the bag-of-words framework, because these two frames share plenty of matched codewords. Assume two codewords $\vv{D}_m^{(k+1)}$ and $\vv{D}_m^{(p+1)}$ are matched due to the fact that (at least) features $f_k$ and $f_p$ are strongly matched with each other. 

If feature $f_k$ is stable across frames $k-1$ and $k$, and it is also matched between these two frames, then the codeword $\vv{D}_m^{(k)}$ generated from $f_k$ for frame $k$ should also match strongly with $\vv{D}_m^{(p+1)}$. If there are enough amount of such stable features across frames $k-1$ and $k$, a loop-closure hypothesis should also be triggered by matching frames $p+1$ with frame $k$. This results in at least two \textit{consecutive} frame indices existing in the hypothesis list. 

Therefore, we impose a temporal constraint on the loop-closure hypotheses generated for a frame: 

\textit{A hypothesis with frame $k$ is accepted, if and only if either frame $k-1$ or frame $k+1$ is also retrieved as a hypothesis}.

\begin{figure}[t!]
  \centering
  \includegraphics[width=1.01\linewidth]{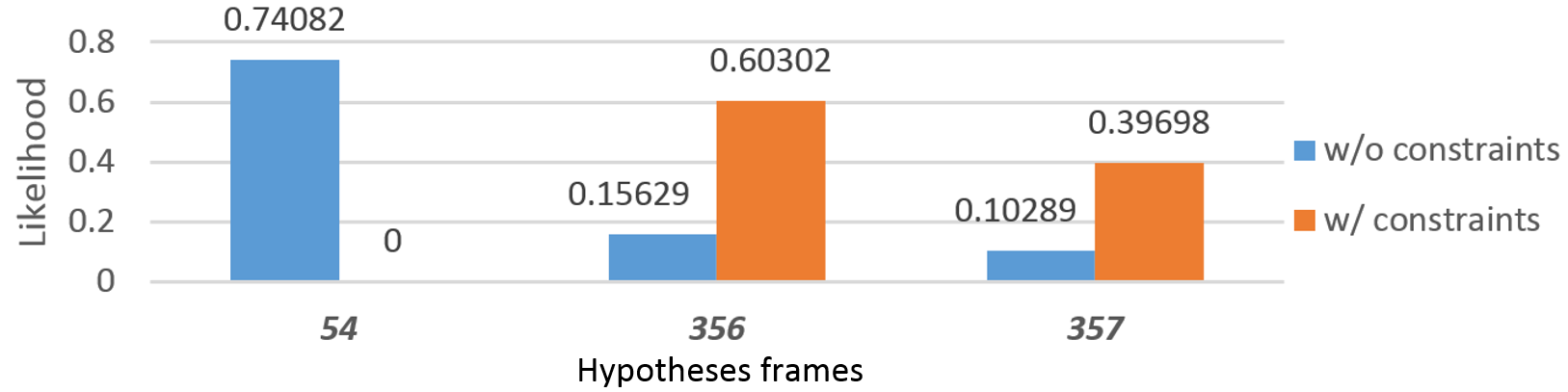}
  \caption{Temporal constraints rejects the false positive hypothesis frame 54. In this example, the true loop-closure frame is 356.   \label{fig:lcTemporalConstraintResult}}
\end{figure}

The temporal constraint on hypotheses reduces the false positive rate and leads to improved detection precision. Fig.\ref{fig:lcTemporalConstraintResult} illustrates an example of using the temporal constraint. True loop-closure exists in frame 356. The hypothesis list \textit{without} temporal constraints includes frame 54 (with likelihood 0.74), frame 356 (with likelihood 0.16), and frame 357 (with likelihood 0.10). The temporal constraint rejects frame 54 (since neither frame 53 nor 55 are in the hypothesis list), the false positive hypothesis. After renormalization, the temporally constrained hypotheses become frame 356 (with likelihood 0.60) and frame 357 (with likelihood 0.40). Frame 356 is therefore retrieved as the final loop-closure index.

\section{Loop Closure Detection System}
\begin{figure}[t!]
  \centering
  \includegraphics[width=0.85\linewidth]{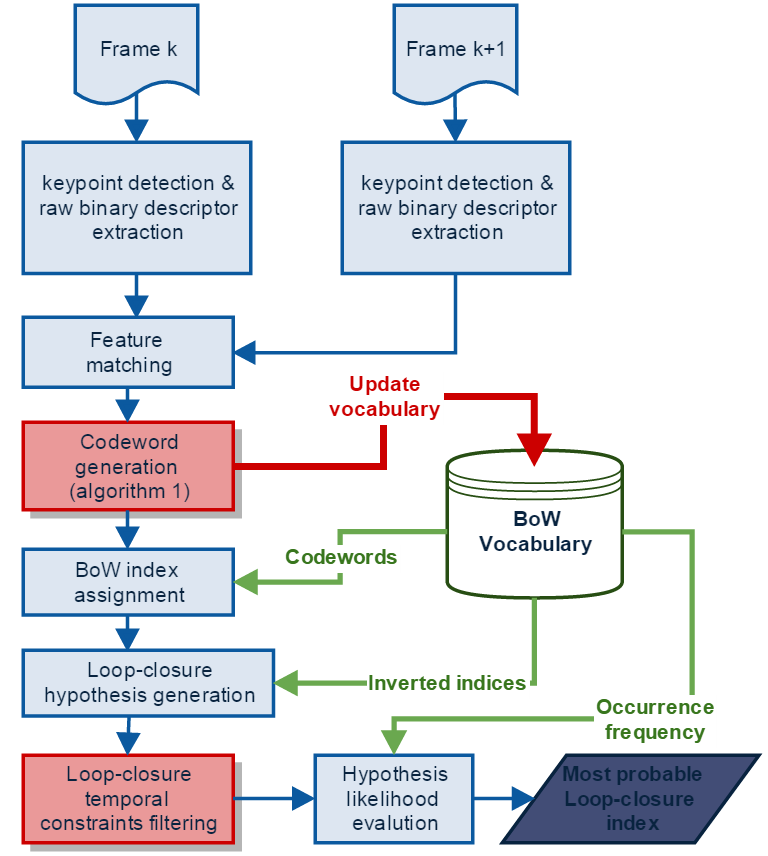}
  \caption{Diagram of the final loop-closure detection system. \label{fig:System_diagram}}
\end{figure}

Our final system design is based the state-of-the-art system IBuILD~\cite{ibuild}, but with the integration of the proposed algorithm, as depicted in Fig.~\ref{fig:System_diagram}. The system is an incremental bag-of-word system without any prior offline training process. Here we discuss some details of the system.

\begin{itemize}
\item The initial feature matching is based on raw binary descriptors extracted from visual keypoints (FAST is used). The raw binary descriptors are from the same binary tests which are used from the codeword learning. A local search on the next frame is then performed to find the best matched features. This step would come for free in an actual visual SLAM system.
\item Masked Hamming distance is used for distance evaluation in all modules except for the initial feature matching.
\item The codeword generation is based on Algorithm~\ref{alg:Codeword_generation}\;. In the actual implementation, since $\vv{x}_1$ and $\vv{x}_2$ are already known from the previous steps, they are directly used to generate the mask $\vv{y}_m$ instead of the raw image patches $I_1$, $I_2$.
\item A codeword merging step is performed when the learned codewords are used to update the vocabulary. This step will iterate through all the codewords generated for the current frame and find the codeword pairs within matching thresholds. These pairs will then be merged according to Algorithm~\ref{alg:Codeword_generation} but treating the two codewords to merge as $\vv{D}_1$ and $\vv{D}_2$. Contrast this merge to~\cite{ibuild} which takes the ``numerical centroid''. For two binary vectors, the ``numerical centroid'' is equivalent to the bit-wise \textrm{OR} operation.
\item The temporal constraint discussed in Section~\ref{subsec:temporal_constraints} filters the hypothesis list before the loop-closure likelihood calculations. It is performed with a single-pass scan on the hypothesis list.
\item The hypothesis with the highest likelihood is output as the final loop-closure index. We perform temporal consistency check of $k=2$ (like~\cite{DBoW}, Section VI.C). 
\end{itemize}

\section{Evaluation}

\begin{figure}[t]
  \centering
  \includegraphics[width=0.95\linewidth]{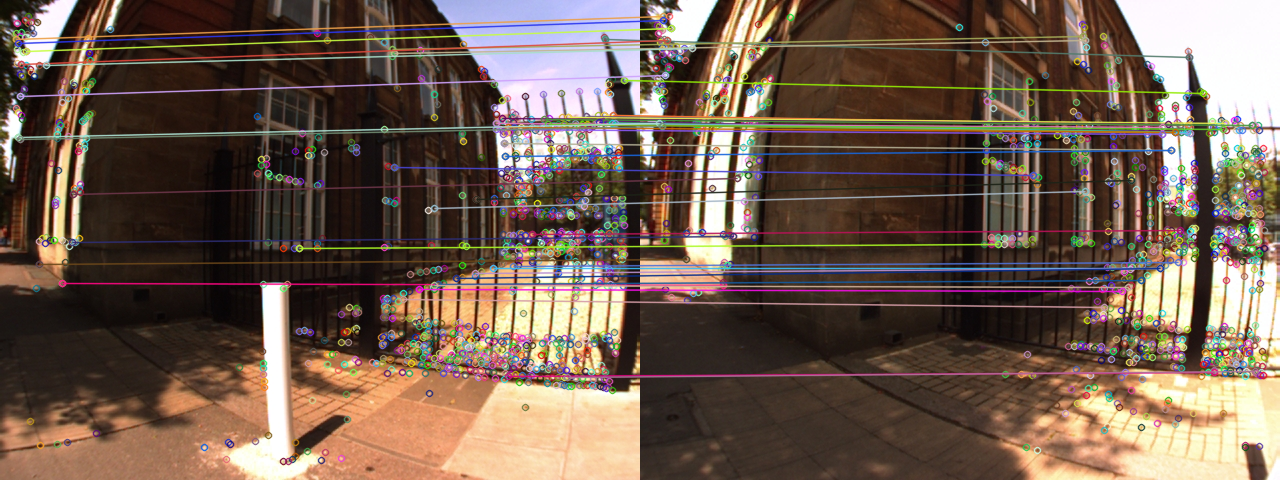}
  \caption{Example frames in \texttt{CityCentre} data set.\label{fig:Example_CityCenter}}
\end{figure}
\begin{figure}[t]
  \centering
  \includegraphics[width=0.95\linewidth]{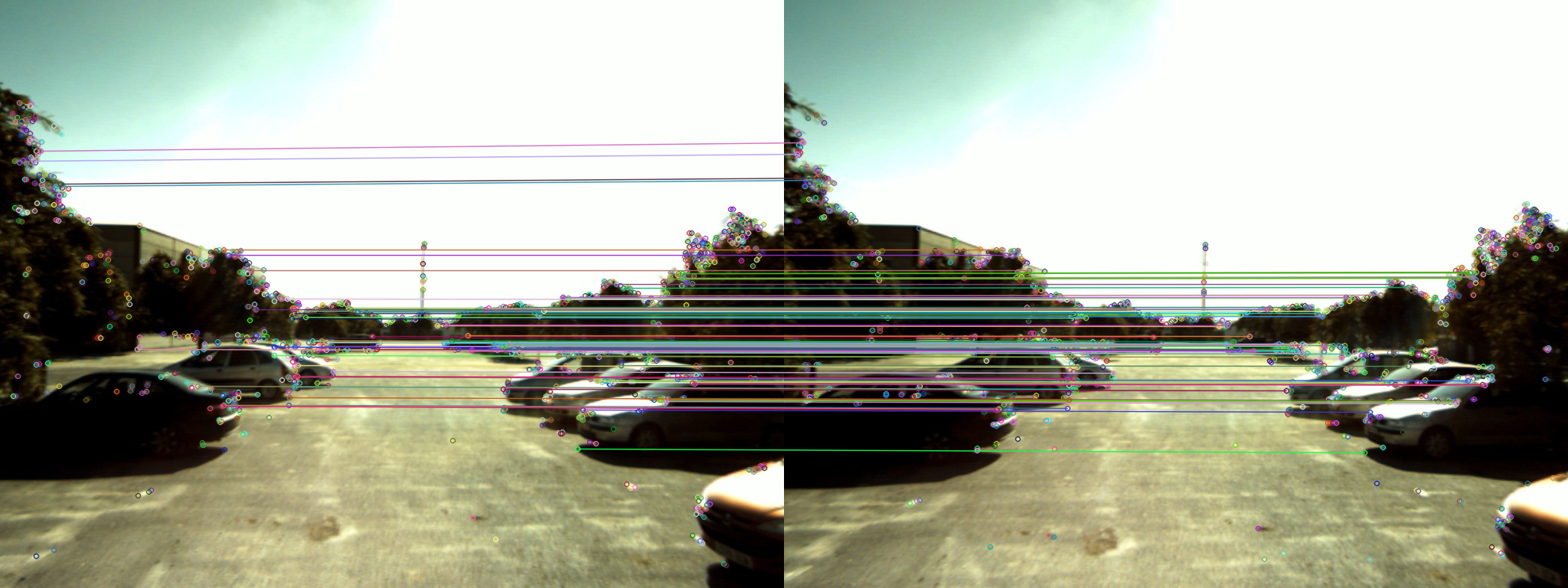}
  \caption{Example frames in \texttt{Malaga09 6L} data set.\label{fig:Example_Malaga}}
\end{figure}
\begin{figure}[t]
  \centering
  \includegraphics[width=0.95\linewidth]{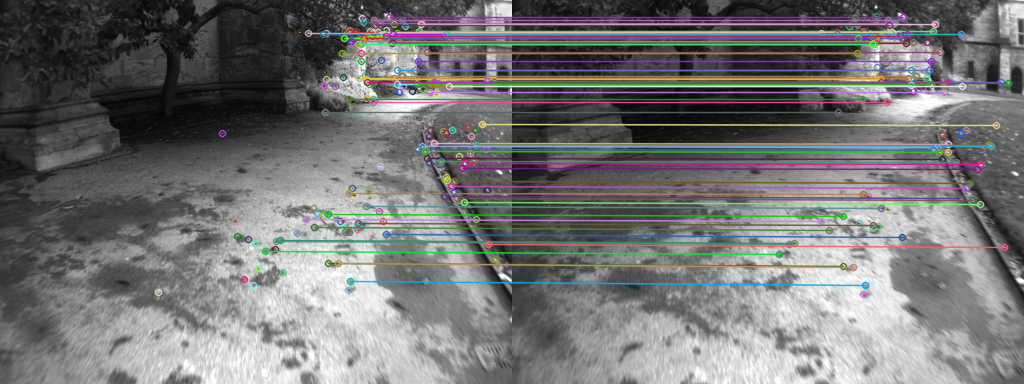}
  \caption{Example frames in \texttt{New College} data set.\label{fig:Example_NewCollege}}
\end{figure}
\begin{figure}[t]
  \centering
  \includegraphics[width=0.41\linewidth]{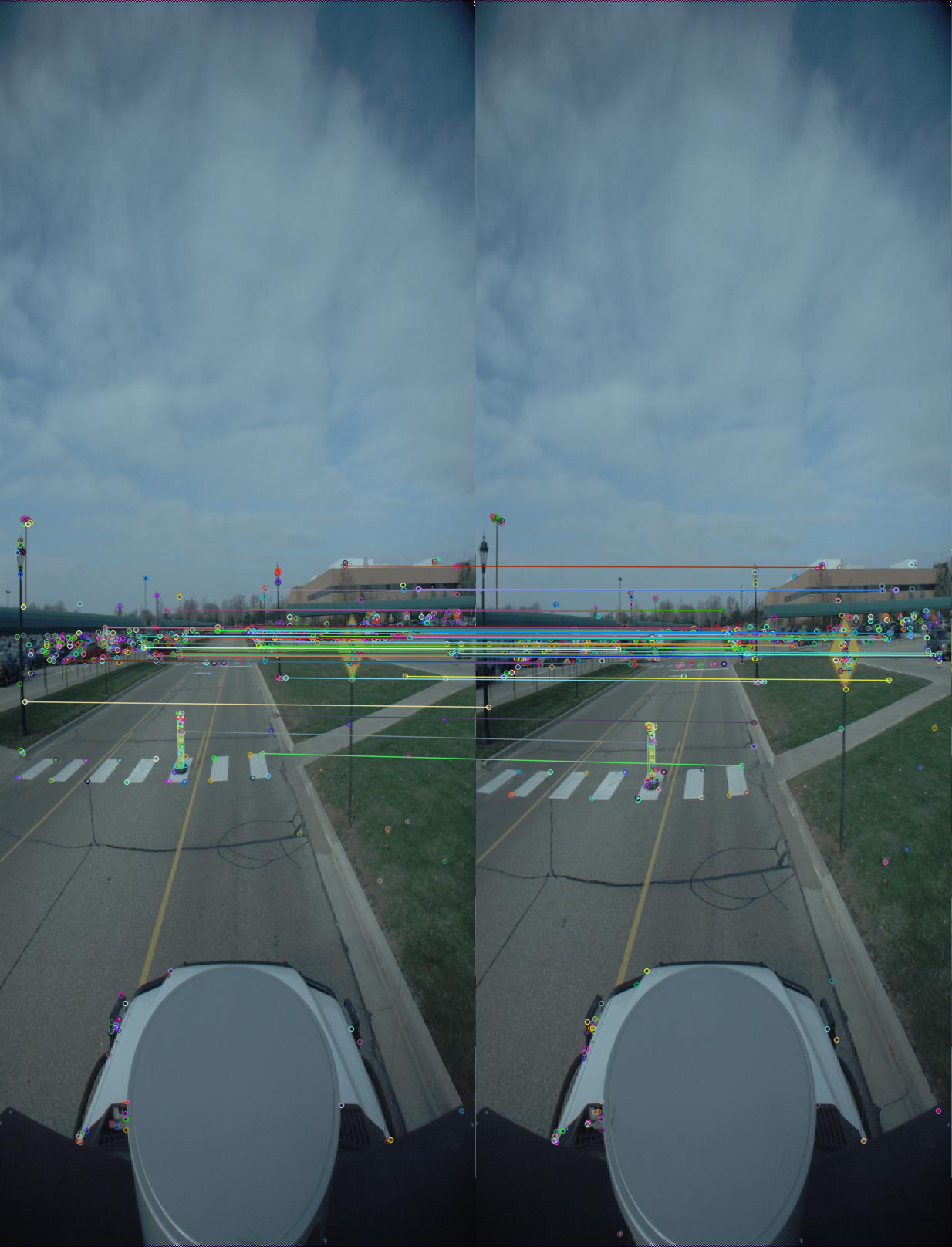}
  \caption{Example frames in \texttt{Ford Campus 2} data set.\label{fig:Example_Ford2}}
\end{figure}

Evaluation used benchmark datasets. To get reasonable parameters, we first performed a coarse-grain parameter sweep on the \texttt{CityCentre} set. The experiments on the other data sets involved modifying the matching threshold only. Performance is compared to three other methods: Fab-Map 2.0~\cite{FabMap2}, Bag-of-binary-words~\cite{DBoW}, and IBuILD~\cite{ibuild}. The machine used was
a Linux desktop with Intel Core i5 quadcore 2.8GHz CPU and 8 GB memory.

\subsection{Datasets and evaluation tool}
The data sets used include four challenging sets:
\texttt{CityCentre}~\cite{FabMap}, \texttt{Malaga09 6L}~\cite{Malaga},
\texttt{New College}~\cite{NewCollege}, \texttt{Ford Campus 2}~\cite{Ford2}.
Table~\ref{Tab:dataset} summarizes their characteristics.
For benchmarking, the evaluation scripts and ground truth files from the
authors of~\cite{DBoW} are used. 

Figures~\ref{fig:Example_CityCenter} to~\ref{fig:Example_Ford2} visualize some matched feature pairs that are used in learning codewords. Notice that in the \texttt{Ford Campus 2} set, part of the vehicle is always visible. We kept only the paired keypoints with pixel coordinate $v \leq 1200$.

\begin{table*}[t]
\centering
\caption{Datasets used for evaluation}
\label{Tab:dataset}
\begin{tabular}{l|l|c|c|c|c}
	\toprule
	{\bf Dataset} & {\bf Environment} & {\bf \begin{tabular}[c]{@{}c@{}}Distance\\ (m)\end{tabular}} & {\bf \begin{tabular}[c]{@{}c@{}}Sensor\\ position\end{tabular}} & {\bf \begin{tabular}[c]{@{}l@{}}Image\\ resolution\end{tabular}} & {\bf \begin{tabular}[c]{@{}c@{}}Number \\ of frames\end{tabular}} \\ \midrule
	CityCentre~\cite{FabMap} & Outdoor, urban, dynamic & 2025 & Lateral & $640 \times 480$ & 2474 \footnotesize{(Left and right cameras)} \\ \hline
	Malaga09 6L~\cite{Malaga} & Outdoor,  slightly dynamic & 1192 & Frontal & $1024 \times 768$ & 869 \\ \hline
	New College~\cite{NewCollege} & Outdoor, dynamic & 2260 & Frontal & $512 \times 384$ & 5266 \\ \hline
	Ford Campus 2~\cite{Ford2} & Outdoor, urban, slightly dynamic & 4004 & Frontal & $600 \times 1600$ & 1182 \\ \bottomrule
\end{tabular}
\end{table*}

\subsection{Experiment and parameter sweep on CityCentre set}
We chose \texttt{CityCentre} set for a coarse-grain parameter sweep because it is a difficult dataset to get high recall with 100\% precision, as will be shown in Table~\ref{tab:PR}\;. The dimensions $L$ of tests and masks are fixed to be 512. The parameter sweep evaluated four parameters: (1) matching threshold $\varPsi$ for $d_{\rm MH}$, (2) keypoints detection threshold $\varUpsilon$, (3) maximum number of matched pairs allowed $\varGamma$, and (4) number of local frames excluded $T_{\rm local}$. Among these parameters, we observed $T_{\rm local}$ can cover a large range, i.e. 20--50, with little change in the results. Moreover, $\varGamma$ mainly controls the size/growth of vocabulary by limiting the accepted pairs of initially matched features. Since it trades off efficiency and precision/recall, we set the value to 100.

\begin{figure}[t!]
  \centering
  \includegraphics[width=0.99\linewidth]{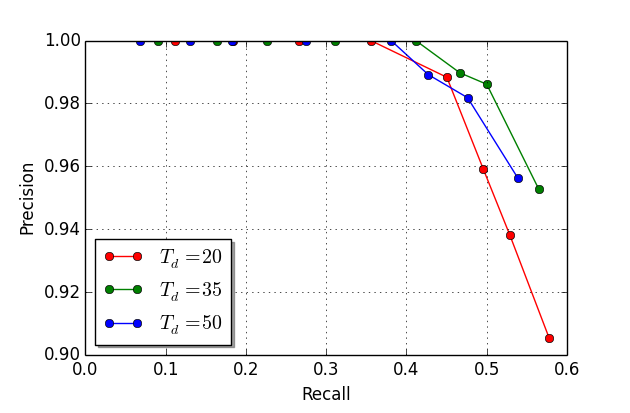}
  \caption{The precision-recall under different detection thresholds on \texttt{CityCentre} set. Each precision-recall curve of a certain detection threshold is plotted by changing the matching threshold for the masked Hamming distances.\label{fig:PR_curve_CityCentre}}
\end{figure}

The most important parameters are matching threshold $\varPsi$ and detection threshold $\varUpsilon$. Figure~\ref{fig:PR_curve_CityCentre} plots the precision-recall curves with $\varPsi = [8, 10, 12, 15, 18, 20, 22, 25]$ under $\varUpsilon = [20, 35, 50]$. The best recall with 100\% precision happens when $\varPsi = 18, \varUpsilon = 35$, as listed in Table~\ref{tab:BestPR_CityCenter}. 

\begin{table}[]
{\centering
\caption{The parameter set which gives the best recall with \textbf{100\% precision} on \texttt{CityCentre} set.}
\label{tab:BestPR_CityCenter}
\begin{tabular}{|l|c|}
\hline
{\bf Parameters} & {\bf Values} \\ \hline
Matching threshold for $d_{\rm MH}$, $\varPsi$ & 18 \\ \hline
Keypoints detection threshold, $\varUpsilon$  & 35 \\ \hline
Maximum number of matched pairs allowed, $\varGamma$  & 100 \\ \hline
Binary test dimensions, $L$ & 512 \\ \hline
Number of local frames excluded, $T_{\rm local}$ & 20 \\ \hline
\end{tabular}}
\end{table}

\subsection{Experiments on all four datasets}

\begin{figure}[t!]
  \centering
  \includegraphics[width=0.99\linewidth]{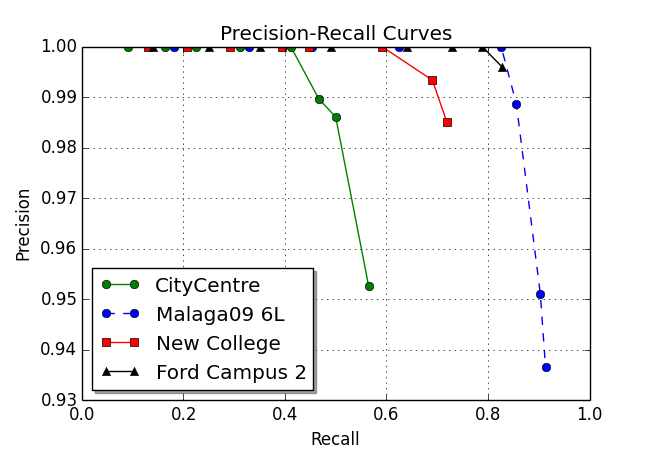}
  \caption{The precision-recall on different data sets using the proposed approach.\label{fig:PR_curve_All}}
\end{figure}

In these experiments, the precision and recall are evaluated by changing $\varPsi$, but keeping $\varUpsilon = 35, \varGamma = 100, T_{\rm local} =20$.
The precision-recall curves of our approach is depicted in Figure~\ref{fig:PR_curve_All}.

Finally, a comparison is provided of our approach to the other approaches~\cite{FabMap2, DBoW, ibuild}, focusing on the best recall under 100\% precision. The results of other approaches are directly from the referred publications. In particular, for bag-of-binary-words method, the \texttt{New College} and \texttt{Ford Campus 2} datasets are used as the training sets with parameter searching, while \texttt{CityCentre} and \texttt{Malaga09 6L} are used as testing set with fixed parameters. For Fab-Map 2.0 on \texttt{Malaga09 6L}, only 462 images are used~\cite{DBoW}. It can be observed that except for the \texttt{Ford Campus 2}, our approach has the highest recalls under 100\% precision.

\begin{table*}
\centering
\caption{(used 462 images)}
\label{tab:PR}
\begin{tabular}{c||c|c|c|c}
\toprule
\multirow{2}{*}{{\bf \emph{Dataset}}} & \multicolumn{4}{c}{{\bf \emph{Performance (precision/recall) of different approaches}}} \\ \cline{2-5} 
 & {\bf Fab-map 2.0~\cite{FabMap2}} & {\bf \begin{tabular}[c]{@{}c@{}}Bag of binary\\ words~\cite{DBoW}\end{tabular}} & {\bf IBuILD~\cite{ibuild}} & {\bf Ours} \\ \hline \hline
{\bf CityCentre} & 100\% / 38.77\%  & 100\% / 30.61\% & 100\% / 38.92\% & 100\% / 41.18\% \\ \hline
{\bf Malaga09 6L} & 100\% / 68.52\% & 100\% / 74.75\% & 100\% / 78.13\% & 100\% / 82.61\% \\ \hline
{\bf New College} & Not available & 100\% / 55.92\% & Not available & 100\% / 59.20\% \\ \hline
{\bf Ford Campus 2} & Not available & 100\% / 79.45\% & Not available & 100\% / 78.92\% \\ \bottomrule
\end{tabular}
\end{table*}

\subsection{Timing}
We collected the timing statistics for learning a codeword as in Algorithm~\ref{alg:Codeword_generation} from \texttt{CityCentre} experiments. 
The timing result under normal system process priority is listed in Table~\ref{tab:timing}, with demonstrates a high efficiency and stability.
In our implementation, the bit-wise operations are handled using C++ \texttt{std::transform} function with bit operation structure (e.g. \texttt{std::bit\_xor}) for \texttt{uchar}. The number of 1s in a binary vector is counted directly using a look-up table indexed by \texttt{uchar} values.

\begin{table}[t]
\centering
\caption{Time (in $\mathbf{10^{-6}}$ sec.) used for learning one codeword using Algorithm~\ref{alg:Codeword_generation}.}
\label{tab:timing}
\begin{tabular}{l|c|c|c|c}
\toprule
{\bf Stat.} & Mean & Standard Dev. & Min & Max \\ \hline
{\bf Time Used} & 14.60 & 0.76 & 13.10 & 16.04 \\ \bottomrule
\end{tabular}
\end{table}

\section{Conclusion}
This work described a method to learn binary codewords online for loop-closure detection. The codewords are learned efficiently in an LDA fashion from matched feature pairs in two consecutive frames, such that the learned codewords encode temporal perspective invariance from the observed motion dynamics. The geometric properties of the learned codewords are mathematically justified. The temporal consistency from the nature of learned codewords is further exploited to cull loop-closure hypotheses. The final incremental system is evaluated with precision/recall and timing results and demonstrate the effectiveness and efficiency of the approach. 

\section*{ACKNOWLEDGMENT}
We'd like to thank Summer Undergraduate Research in Engineering (SURE) grant (NSF award number: EEC-1263049) for supporting Mason Lilly. We sincerely thank the authors of \cite{ibuild} for sharing the main components of his IBuILD implementation, and thank the authors of \cite{DBoW} for providing their evaluation scripts and ground truth.


\end{document}